\documentclass[letterpaper, 10 pt, conference]{ieeeconf}  

\usepackage{graphicx}
\usepackage{flushend}
\usepackage{balance}
\usepackage{xcolor}

\IEEEoverridecommandlockouts                              

\title{\LARGE \bf
Threading the Needle: Test and Evaluation of Early Stage UAS Capabilities to Autonomously Navigate GPS-Denied Environments in the DARPA Fast Lightweight Autonomy (FLA) Program
}

\author{Adam Norton$^{1}$ and Holly A. Yanco$^{1}$
\thanks{*This material is based upon work suppoted by DARPA and SPAWARSYSCEN Atlantic under Contract No. N65236-16-C-8002. Any opinions, findings and conclusions or recommendations expressed in this material are those of the authors and do not necessarily reflect the views of DARPA or SPAWARSYSCEN Atlantic. Distribution Statement ``A'' (Approved for Public Release, Distribution Unlimited).}
\thanks{$^{1}$New England Robotics Validation and Experimentation (NERVE) Center, University of Massachusetts Lowell, Lowell, MA, USA
        {\tt\small adam\_norton,holly\_yanco@uml.edu}}%
}

\begin{document}

\maketitle
\thispagestyle{empty}
\pagestyle{empty}

\begin{abstract}

The DARPA Fast Lightweight Autonomy (FLA) program (2015--2018) served as a significant milestone in the development of UAS, particularly for autonomous navigation through unknown GPS-denied environments.
Three performing teams developed UAS using a common hardware platform, focusing their contributions on autonomy algorithms and sensing.
Several experiments were conducted that spanned indoor and outdoor environments, increasing in complexity over time.
This paper reviews the testing methodology developed in order to benchmark and compare the performance of each team, each of the FLA Phase 1 experiments that were conducted, and a summary of the Phase 1 results.

\end{abstract}

\section{Introduction}

The past 25 years of research and development in aerial robotics has seen tremendous growth in the adoption of systems as well as the advancement of capabilities including increased speed, more reliable autonomy, and powerful onboard computing.
There are several significant milestones in this history that saw the domain shifting from primarily GPS-driven systems to those capable of operating in GPS-denied environments.
One such milestone was the DARPA Fast Lightweight Autonomy (FLA) program~\cite{darpa-fla}, running from 2015--2018, seeking to develop minimalistic algorithms for high-speed autonomous navigation of UAS in cluttered environments without prior knowledge, remote control, or GPS.
Unlike other DARPA funded robotics programs like the Robotics Challenge (DRC) and Subterranean (SubT) Challenge, FLA did not conduct public events or competitions, so visibility of the program was somewhat limited in comparison.
However, the experimentation conducted during the program is a significant benchmark, demonstrating the early stage capabilities of UAS in this burgeoning field and setting the stage for future developments.
This paper presents an overview of the test and evaluation conducted for the DARPA FLA program, reviewing the testing methodology used and a summary of the results.

\section{DARPA Fast Lightweight Autonomy (FLA)}

From the start, FLA had several objectives towards achieving autonomous UAS navigation of complex, urban, cluttered environments, including flying up to 20~m/s up to 1 km of range for a 10-minute mission, without high-quality prior knowledge, without communications back to the operator, and without GPS, all performed onboard the UAS with 20~W of computing power.
The program utilized a wide range of algorithmic and sensing approaches by having performing teams focus on the development of autonomy rather than hardware design.
A common commercial UAS platform (COTS airframe, motors, and propellers, with an open-source autopilot, thrust-to-weight ratio of 2.8, and able to fit through doors and windows) was provided and teams outfitted their own sensors, processors, and software. 
A series of experiments were planned and executed such that teams' solutions could be evaluated and compared in progressively more difficult environments, adding complexity including denser obstacles, longer distances, and varied lighting.

\begin{figure}
\includegraphics[width=\columnwidth]{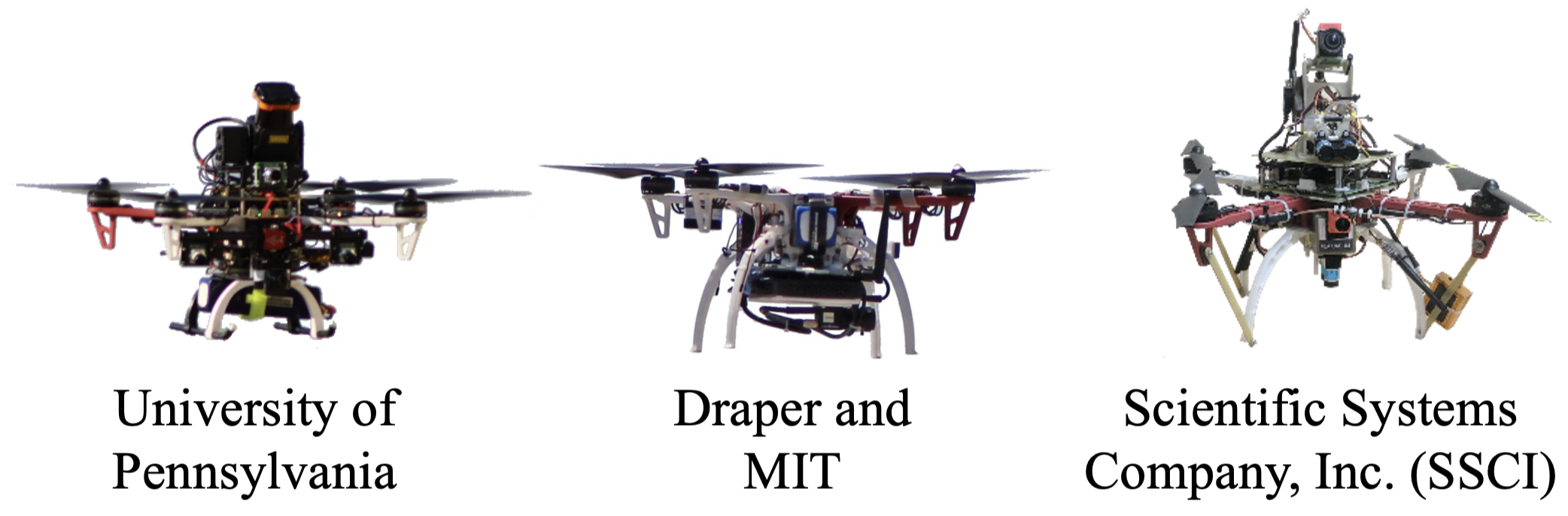}
\caption{The three performing teams of the DARPA FLA program.}
\vspace{-0.5cm}
\label{fig:performers}
\end{figure}

Three performer teams participated in FLA (see Fig.~\ref{fig:performers}).
\textbf{University of Pennsylvania} utilized integrated sensing, planning, and control for fast flight with novel sensors and processing~\cite{mohta2018fast}.
\textbf{Draper and MIT} leveraged symbolic perception techniques for provably-correct motion primitives, including their ``Samwise'' pose and state estimation~\cite{paschall2017fast,steiner2017vision}.
\textbf{Scientific Systems Company, Inc. (SSCI)} leveraged a bio-inspired technique with tightly-coupled perception and control, using expansion rate techniques to determine collision risk from obstacles~\cite{escobar2018r}.
A common strategy for all teams, though, was the use of monocular cameras in order to reduce weight and the required computing, as well as sensors for altitude detection (single point lidar) and inertial sensors for pose measurement.
Some teams also used stereo cameras and spinning lidars for certain experiments; more information on each team's solution can be found in their cited papers.

The DARPA FLA program ran in two phases. 
During Phase 1 (2015--2017), all teams were tasked with performing the same challenges across a series of experiments allowing for direct side-by-side comparison of performance to be evaluated.
For Phase 2 (2017--2018), each team was instructed to focus on their respective strengths, pushing each team to achieve autonomous navigation capabilities across different mission profiles.
Given that the test and evaluation activities for Phase 2 were not designed for side-by-side comparison, this paper only covers the experimentation conducted during Phase 1 of the program.

\section{Testing Methodology}

With the focus of FLA on the development of autonomy for navigation rather than search, the UAS had to be given some form of mission plan to execute.
In more typical navigation scenarios, a mission specification requires at least the location of a goal to be reached, often conveyed as a coordinate either globally as GPS or locally as x,y,z in a given map.
However, given the UAS were operating without GPS, there were limited methods to specify a goal and for the UAS to validate whether or not it had reached that goal.
If a soldier were in a similar scenario, they could rely on the use of paper maps (which may or may not be accurate) using a compass to navigate, looking for a target feature in the environment as their goal point (e.g., building or vehicle).
Following this analogy, several test and evaluation elements were developed and implemented.

\begin{figure}
\includegraphics[width=\columnwidth]{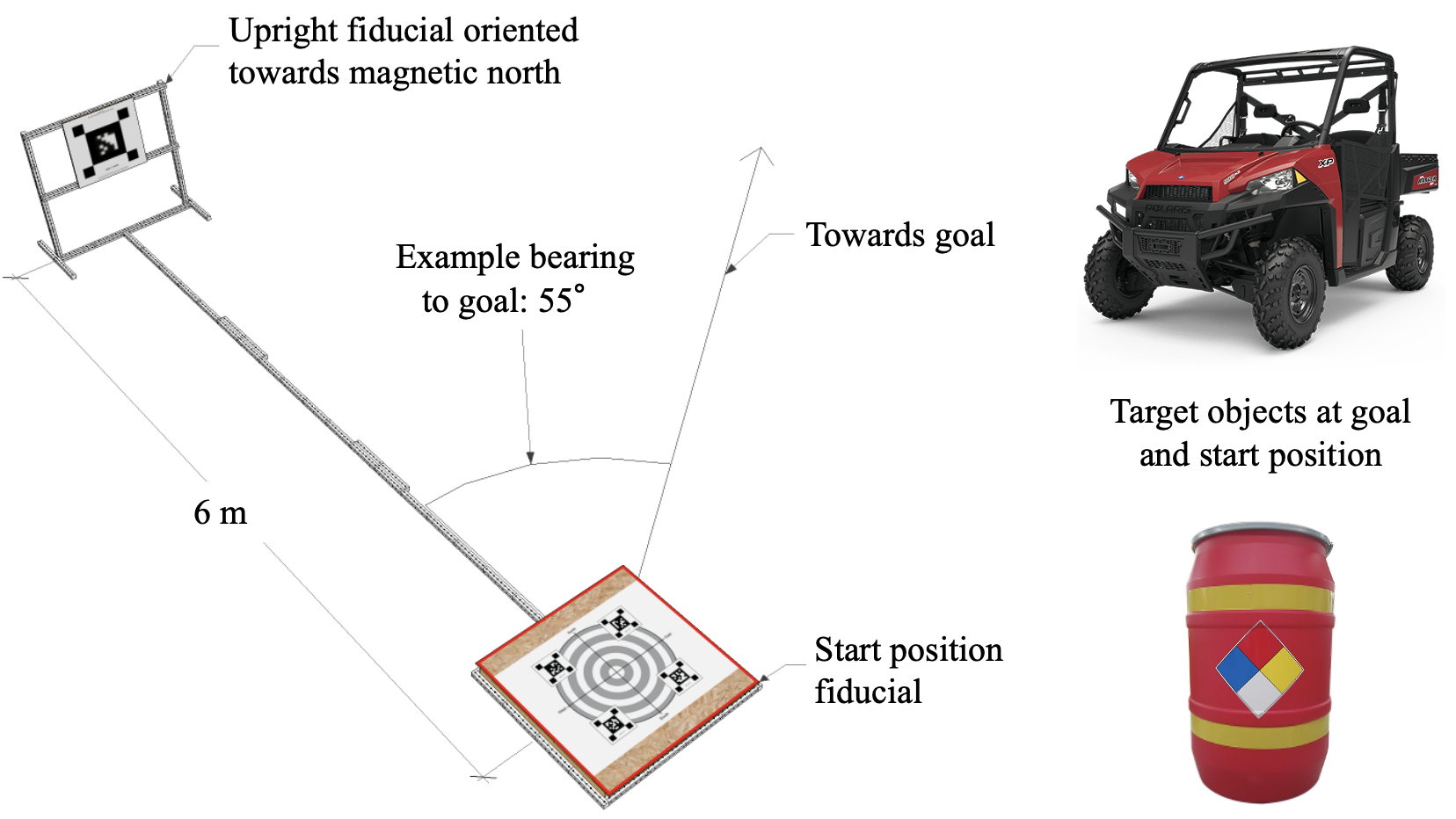}
\caption{Testing artifacts at the start position and goal.}
\vspace{-0.5cm}\label{fig:start-goal}
\end{figure}

\textbf{Start position and goal artifacts.}
Fiducials consisting of AprilTags were placed at the start position that could be used by the UAS to align to magnetic north.
This was accomplished using horizontal fiducials on the takeoff platform for the UAS to identify using a downward-facing camera and a vertical fiducial placed 6 m away that was aligned to magnetic north to be identified using a forward-facing camera.
A red target object -- either a Polaris vehicle or a barrel with hazmat labels -- was placed at the goal and start position which the UAS would seek to recognize to further guide its navigation to the intended location (Fig.~\ref{fig:start-goal}).

\begin{figure}
\includegraphics[width=\columnwidth]{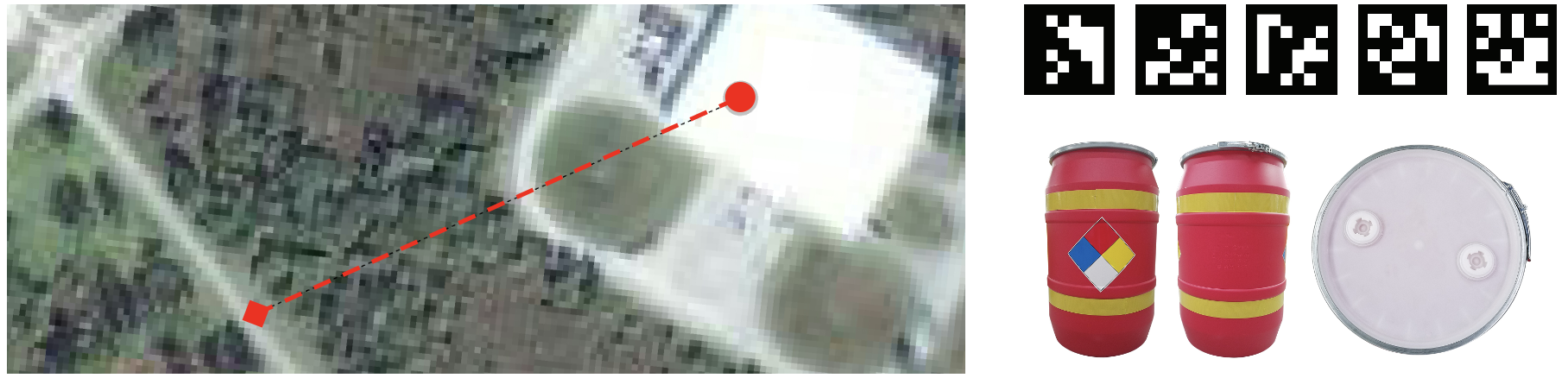}
\caption{Example contents of a mission file: (clockwise from left) overhead map with trajectory, AprilTags, and target object images.} \label{fig:mission-file}
\end{figure}

\textbf{Mission files}.
Provided to each team to use for UAS navigation, mission files (Fig.~\ref{fig:mission-file}) consisted of an XML file with relative distance and bearing from magnetic north between the start position and goal, dimensions and images of the target objects, the AprilTags used in the start position fiducials, satellite imagery of the area (e.g., Google Maps) oriented to magnetic north with a line drawn from the start position to the goal, and elevation limitations to ensure interaction between obstacles and the UAS (e.g., 4.3 m elevation limit ensures the UAS will not fly above a building or tree).
Using these elements, each UAS could align with the fiducials at the start position, rotate in place the relative bearing to the goal, navigate and avoid obstacles along that trajectory, detect the target object at the goal, turn around, and return to start.
Mission files also ensured fairness across teams by giving them all the same level of information.

\begin{figure}
\includegraphics[width=\columnwidth]{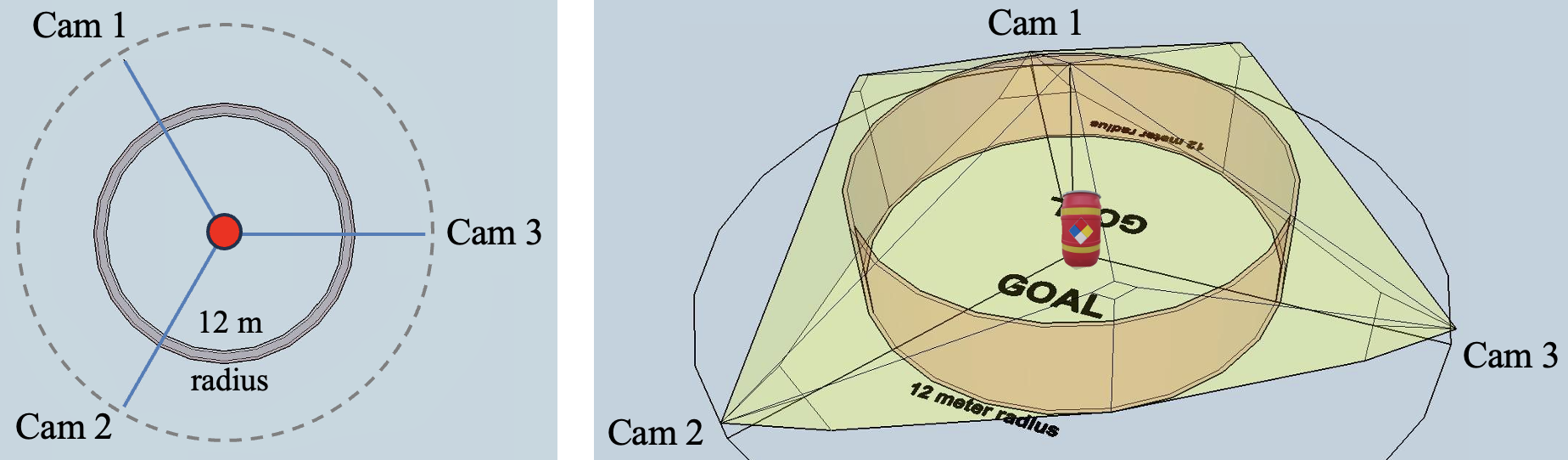}
\caption{Diagrams of the camera convergence method.} 
\vspace{-0.5cm}
\label{fig:cameras}
\end{figure}

\textbf{Successful navigation}.
Criteria was established for the UAS having successfully reached the goal and/or returned to start.
A cylindrical volume with approximately 12 m radius and 5 m height around each target object was established as the acceptable level of navigation accuracy.
Three time synchronized cameras were positioned radially around the target object such that the combined fields of view from each camera covered the cylindrical volume, meaning that if the UAS was observed in the frames of all cameras then it was physically within the acceptable threshold.
This was dubbed the ``camera convergence method'' (Fig.~\ref{fig:cameras}).

\textbf{Metrics}.
There were two primary metrics of performance. \textit{Navigation efficacy} was evaluated in two segments: reaching goal and returning to start. 
\textit{Average speed} (m/s) was evaluated by dividing the length of the approximate shortest path from start to goal and back while navigating around obstacles by the navigation time; this was only evaluated if the goal was reached (for the first segment; ``halfway'') or if the UAS returned to start (for both segments combined; ``roundtrip'').

\textbf{Human baseline}.
Lastly, as a point of comparison, baseline performance metrics were collected for each task where a pilot teleoperated a UAS made from the same common test platform with 1.4 kg of weight added to be on par with the performing teams.

\section{Experiments}

\begin{figure*}
\centering\includegraphics[width=0.95\textwidth]{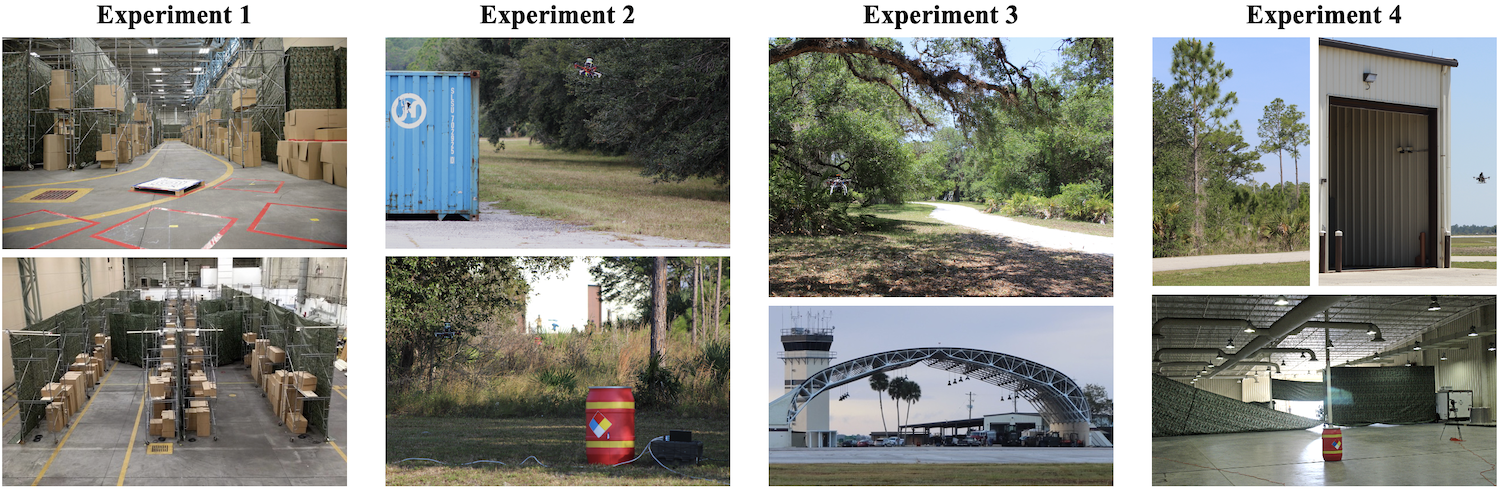}
\caption{Experiments conducted during Phase 1 of the DARPA FLA program.} 
\vspace{-0.3cm}
\label{fig:experiments}
\end{figure*}

\begin{figure*}[t!]
\centering\includegraphics[width=0.95\textwidth]{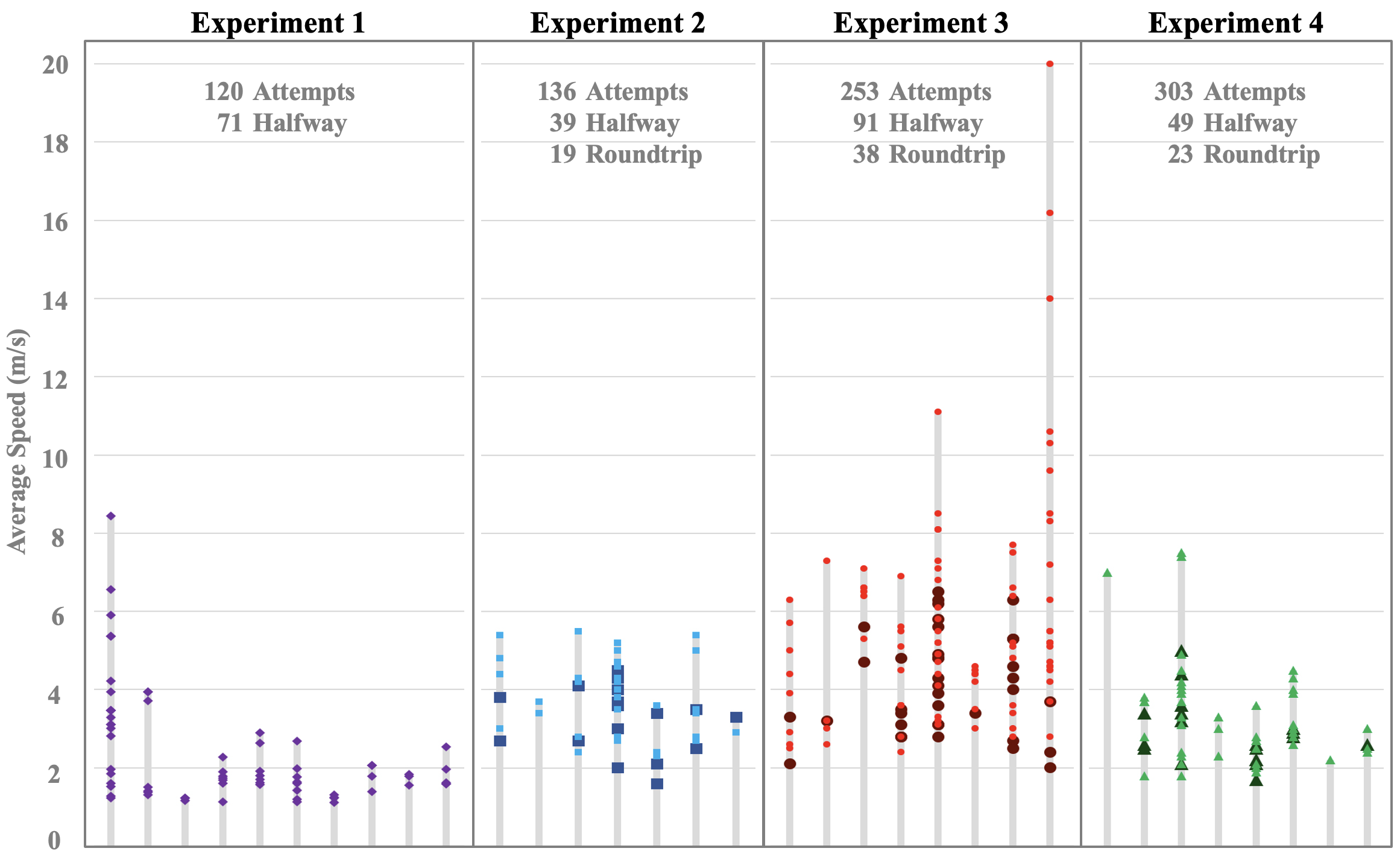}
\caption{Anonymized results for DARPA FLA Phase 1. Colors and shapes used correspond to each experiment: small symbol = successfully reached goal; large symbol = successfully returned to start. Each bar corresponds to a unique task design used during that experiment (e.g., 7 tasks in experiment 2). } 
\vspace{-0.5cm}
\label{fig:results}
\end{figure*}

During Phase 1 of FLA, four experiments were conducted with each adding new challenges progressing from (1) indoor warehouse environments, (2) outdoor natural terrain, (3) outdoor natural terrain with structures and lighting transition, and (4) indoor and outdoor terrain with transitions between the two (Fig.~\ref{fig:experiments}).
Anonymized results from all experiments are provided in Fig.~\ref{fig:results}.

\subsection{Warehouse (Experiment 1)}

\begin{figure}
\includegraphics[width=\columnwidth]{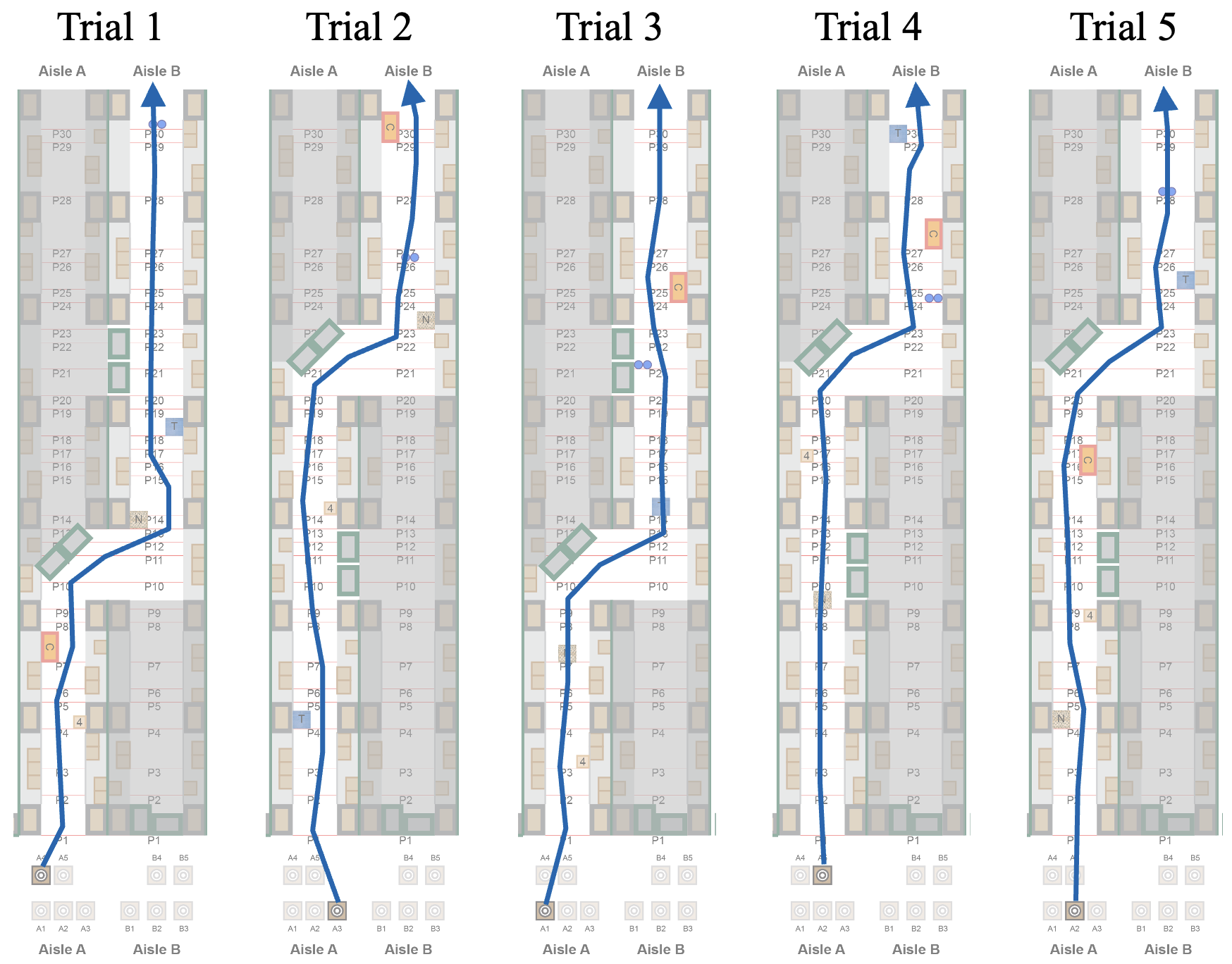}
\caption{Example of variable task designs across trials.} 
\vspace{-0.5cm}
\label{fig:variable-tasks}
\end{figure}

Experiment 1 was held in April 2016 in Massachusetts where a warehouse test course was fabricated inside of a hangar made out of rolling scaffolding towers, tarps, and cardboard boxes.
The warehouse test course consisted of two tight aisles defined by racking and boxes as well as several box obstacles strewn throughout.
Given the early stage of the FLA program, teams were only tasked with reaching the goal. 
In order to exercise the robustness of their autonomy and prevent overfitting of their algorithms, the test course was designed to allow for multiple variations of a task design to be implemented by moving obstacles to different locations.
Start positions were also varied between trials.
See Fig.~\ref{fig:variable-tasks} for an example.
Given that all tasks took place entirely indoors, the overhead imagery provided in the mission files for this experiment were less useful for the teams.
Instead, teams were provided specifications on the possible obstacles used throughout the test course, minimum and maximum dimensions of aisle openings, and text descriptions of the task layout.
For example, the task design shown in Fig.~\ref{fig:variable-tasks} was described as ``double aisle, lateral movement between aisles (1 forced aisle change), 45° turn, with obstacles throughout requiring positive elevation changes and lateral movements.''
Across teams, 59\% of attempted runs (71/120) were successful with average speeds up to 8.5 m/s.

\subsection{Natural Terrain (Experiments 2 and 3)}

Moving outdoors, Experiment 2 took place in November 2016, with task designs that included natural terrain with large open areas, dense tree lines, and a shipping container or building to avoid.
Unlike Experiment 1, no variation was induced across trials for a given task.
Given the dynamic environmental factors including lighting and wind, such induced task layout variation was determined to not be required. 
Teams were now tasked with navigating ``roundtrip'' by returning to start after reaching the goal. 
The addition of natural vegetation and complex scenes made for very challenging tasks, with 29\% of runs (39/136) achieving ``halfway'' success (reaching goal only) and only 14\% returning to start (19/136), with average speeds up to 5.5 m/s.

Experiment 3 took place in January 2017 and added more complex natural environments as well as more manmade structures to navigate around and through (see Fig.~\ref{fig:experiments} for an example of a hangar tent that was used). 
Task designs involving these structures and requiring the UAS to navigate under tree canopy made for more dramatic lighting changes.
Many more attempts were made compared to Experiment 2, with overall success increasing to 36\% (91/253) for reaching goal and 15\% (38/253) for subsequently returning to start.
Most notably, though, is achieving the program goal of autonomous flight at 20 m/s.

\subsection{Indoor and Outdoor Terrain (Experiment 4)}

The final test event of Phase 1 -- Experiment 4 -- took place in May 2017 and used task designs that spanned the full FLA mission profile: starting in outdoor natural terrain, into a warehouse for indoor navigation, and then back outside to return to start.
With the widest variety of environments to date, this proved to be the most challenging to the teams. 
In particular, the final task design of the event, which required the UAS to travel through dense scrub vegetation, across an open tarmac, through a large roll up door, traverse through a complex indoor environment to reach goal, and back to return to start.
Each of the prior experiments utilized elements of this full mission, with the addition of more substantial lighting transitions when moving from the bright outdoor sun to a somewhat dimly lit interior and back.
Across the runs performed by all teams, 16\% success (49/303) was achieved for reaching the goal, while only 8\% (23/303) successfully returned to start, with average speeds up to 7.5 m/s.

\section{Conclusion}

Overall, the advances made across the DARPA FLA program represent a significant milestone in the development of UAS navigation in GPS-denied environments.
While the actual use case of FLA has not yet been fully realized in commercial platforms (i.e., autonomous navigation in GPS-denied environments relying only on magnetic north and satellite imagery), its advancements in research have continued.
For example, visual inertial odometry (VIO) solutions are now essentially ubiquitous on UAS, with many systems relying primarily on 2D camera data to navigate and avoid obstacles.
Prevalent commercial UAS like those from Skydio use similar techniques to avoid obstacles during teleoperation and autonomous flights, with maximum speeds of up to 20 m/s~\cite{skydio}, which was another FLA program goal.
Autonomous navigation to explore and map unknown indoor environments is the primary application for systems like the Shield AI Nova 2~\cite{shieldai} and is performed primarily using RGBD cameras.
More recent programs like the DARPA Subterranean Challenge~\cite{chung2023into}, while not strictly focused on UAS, did involve many UAS platforms as part of teams' solutions to navigate and map unknown underground environments.
These are just a few examples of the aerial robotics field carrying the torch of FLA forward, advancing from the high-risk and frail autonomy that ``threaded the needle'' at the time, albeit with one eye open (i.e., monocular camera).

\bibliographystyle{ieeetr}
\bibliography{references}

\end{document}